\documentclass[10pt,twocolumn,letterpaper]{article}

\usepackage{wacv}              

\usepackage{multirow}

\definecolor{wacvblue}{rgb}{0.21,0.49,0.74}
\usepackage[pagebackref,breaklinks,colorlinks,allcolors=wacvblue]{hyperref}

\title{Scoring, Not Learning: Decomposing Catalog-Consolidation Gains in Web-Scraped Retail Product Recognition}

\author{Kaden Leis, Xiang Ma\\
University of Wisconsin-Eau Claire
}

\begin{document}
\maketitle

\begin{abstract}
Retail product catalogs assembled from web image search are cheap to build and pervasively noisy: queries return packaging shots instead of products, and entries proliferate under names that cannot be distinguished visually. Consolidating such a catalog can look like an alternative to scaling the model. We show this is not a comparison between two ways of improving a recognizer, but between improving a recognizer and changing what counts as correct.

We study a public archive of $165{,}769$ cheese images across $1{,}824$ query-named entries. A packaging filter and near-duplicate removal reduce it to a $998$-class recognition set on one fixed, group-aware split. Canonicalizing entries to their base type ($844$ classes) raises top-1 accuracy by $4.0$ points. Scoring the unchanged model against the merged label space, by summing its predicted probability within each canonical group, recovers $+4.27$ points with no retraining ($+4.25$ over six seeds). Retraining on the consolidated catalog then adds $-0.10\pm0.53$, $+0.08\pm0.65$ and $-0.10\pm0.65$ points at $28$, $50$ and $88$\,M parameters, indistinguishable from zero at every capacity ($\pm$ is the across-seed standard deviation). That component is specific, not mechanical: random merges with the identical size profile yield only $+0.68$ points, and the three backbones agree to within $0.1$ points. But it is overwhelmingly category coarsening, not alias correction: across the $37$ merged groups, type coarsenings supply $94\%$ of the gain and alias collapses only $7\%$.

We also detect no interaction between consolidation and backbone size: tripling backbone parameters is worth $+1.50\pm0.57$ points on the raw catalog and $+1.42\pm0.44$ on the canonical one, a within-seed interaction of $-0.08$ $[-0.79,+0.63]$. On this archive, the reported consolidation gain primarily reflects the evaluation label space, and separating that out costs one evaluation pass.

\end{abstract}

\section{Introduction}
\label{sec:intro}

Recognizing individual products is a core capability for physical retail. Shelf auditing, self-checkout, planogram compliance and catalog maintenance all reduce to deciding which of many visually similar items an image depicts. Curated retail benchmarks exist~\cite{goldman2019sku110k, peng2020rp2k, bai2020products10k, wei2019rpc, klasson2019grocery}, but each covers a small slice of a real assortment, and extending coverage is expensive. Issuing product names as web-search queries is the standard shortcut.

This approach is cheap, but it inherits two problems specific to retail imagery. A query such as \emph{``brie cheese''} returns many photographs of supermarket \emph{packaging} rather than the product, so branding becomes a confound across unrelated items. The label space is also a product catalog rather than a designed taxonomy: \emph{``cheddar''}, \emph{``mature cheddar''} and \emph{``orkney extra mature cheddar''} appear as three query-named classes despite being difficult to separate at web-image resolution. Collapsing entries of this kind to a base type is routine work for anyone maintaining a product database.

We study a public archive of $165{,}769$ readable cheese images in $1{,}824$ query-named folders. It is a deep fine-grained slice of one grocery category, and it exhibits both problems acutely. A third compounds them: a severely long-tailed distribution over those already redundant entries. The images are web photographs rather than shelf or checkout captures, so what transfers to a store deployment is the catalog-side finding, where the label space rather than the imaging condition is the variable.

Data-centric work often reports that improving the data can rival scaling the model~\cite{sorscher2022pruning, gadre2023datacomp, northcutt2021labelerrors}, which would make a catalog-hygiene project an alternative to a larger GPU budget for a retail team. That work, however, keeps the evaluation label space fixed (Section~\ref{sec:related}), whereas catalog consolidation changes it, and that difference hides an asymmetry. Canonicalizing entries to their base type does not merely recover accuracy a redundant label space was suppressing. It also \emph{changes the task}, replacing a $998$-way problem with an easier $844$-way one, whereas scaling the backbone leaves the task fixed. Comparing the two head-to-head therefore mixes a definitional effect with whatever retraining adds. We separate them by scoring a single frozen model under both label spaces, a control that costs one evaluation pass.

We therefore treat consolidation throughout as the adoption of a coarser, predefined label space, and ask what accuracy gain that space provides. We do not treat it as a demonstration that the merged entries name one product, which our name-based map does not establish.

Our contributions are:
\begin{itemize}
\itemsep0.2em
\item A pipeline for turning a noisy web-scraped product archive into a usable recognition set: a packaging filter and a duplicate-group split that removes the pixel-identical train--test overlap a filename split leaves behind.
\item A decomposition of the consolidation gain into a definitional component and a retraining residual, using hierarchical marginalization as the rescoring control it requires. The definitional component accounts for the entire gain within our resolution, the residual is indistinguishable from zero at all three capacities, and random-merge nulls show the component is specific to the merged classes rather than to class-count reduction (Table~\ref{tab:stage3}, Section~\ref{sec:stage3res}).
\item No detectable interaction between curation and capacity over a $3.1\times$ parameter range, with comparisons paired within each of six seeds.
\end{itemize}

\section{Related Work}
\label{sec:related}

\noindent\textbf{Vision-language pretraining.} CLIP~\cite{radford2021clip} learns aligned image-text representations that transfer well downstream. We use a frozen ViT-L/14 encoder~\cite{dosovitskiy2021vit} via OpenCLIP~\cite{ilharco2021openclip}, with the original \texttt{openai} weights, for feature extraction and filtering. Zero-shot prompt matching performs poorly on the rare proper-noun names in our archive, and Section~\ref{sec:stage2res} shows that this poor performance persists after both prompt engineering and the removal of label redundancy, leaving limited text grounding as an untested candidate explanation.

\noindent\textbf{Fine-grained visual categorization.} FGVC benchmarks~\cite{wah2011cub, vanhorn2018inaturalist, bossard2014food101} define explicit category spaces, although supervision can remain noisy: Food-101 deliberately retains noise in its training set. High-capacity backbones such as ConvNeXt~\cite{liu2022convnext} provide a basis for fine-tuning. Chang \etal~\cite{chang2021flamingo} question which label granularity users need, and Bertinetto \etal~\cite{bertinetto2020mistakes} weigh hierarchy-aware error severity. Reading a fine-grained classifier out at a coarser level by summing the predicted probabilities of its child classes is the standard marginalization readout in hierarchical classification~\cite{deng2012hedging, jain2023amendment}; we do not propose this readout but apply it as a control, to separate a change in the evaluation label space from a change in the model. Our label space is itself an artifact of scraping, never designed to be mutually exclusive.

\noindent\textbf{Learning from web supervision and label noise.} WebVision~\cite{li2017webvision} studies learning from noisy web labels, and Krause \etal~\cite{krause2016noisy} show that for fine-grained recognition the volume of noisy web data can outweigh its noisiness. WebFG-496~\cite{sun2021webfg} formalizes weakly supervised fine-grained recognition. This line of work largely treats noise as \emph{instance-level} (wrong image for a correct class). The dominant problem in our archive is \emph{class-level}. Distinct folders are not separable at web-image resolution, whether because they name one product or because they name products of one type, and no per-image relabeling can fix either.

\noindent\textbf{Data-centric learning and dataset curation.} A growing body of work argues that dataset quality, not model scale, limits progress. Label errors reorder model rankings~\cite{northcutt2021labelerrors}, and much of ImageNet's residual error stems from label definitions rather than model failure~\cite{beyer2020imagenet, vasudevan2022bagel}. Relabeling improves fixed architectures~\cite{yun2021relabel}, pruning can improve on power-law data scaling~\cite{sorscher2022pruning}, and semantic deduplication reduces training cost with little accuracy loss~\cite{abbas2023semdedup}. DataComp and MetaCLIP demonstrate gains from data curation under controlled training procedures~\cite{gadre2023datacomp, xu2024metaclip}. Most of these operations leave the evaluation label space fixed. Ours does not, and that is the case we study: when curation changes the label space, part of the measured gain is definitional.

\noindent\textbf{Retail product recognition.} Retail benchmarks cover shelf-object detection~\cite{goldman2019sku110k}, checkout imagery~\cite{wei2019rpc}, SKU-level recognition~\cite{peng2020rp2k, bai2020products10k}, grocery hierarchies~\cite{klasson2019grocery}, instance segmentation~\cite{follmann2018d2s} and grocery-category classification~\cite{jund2016freiburg}. These tasks have deliberately defined annotations and category spaces. Our archive inherits its labels from search queries, so we treat catalog structure as the primary variable.

\noindent\textbf{Long-tailed recognition.} Reweighting the loss toward minority classes raises their accuracy but can reduce overall accuracy under naturally imbalanced test distributions, as we observe. We use inverse-frequency weights $w_c=N/(Cn_c)$, for $N$ training images over $C$ classes with $n_c$ in class $c$, not the effective-number scheme of Cui \etal~\cite{cui2019classbalanced}.

\section{The Archive}
\label{sec:archive}

The dataset derives from the public Cheese Pics collection on Kaggle~\cite{kagglecheese}, which contains web-search images in folders named by cheese. We discard $8$ unreadable files. Table~\ref{tab:dataset} gives its statistics and those of every subset below. We contribute no new images: the contribution is the filtering, reorganization and evaluation of an existing archive.

\begin{figure}[t]
\centering
\includegraphics[width=\columnwidth]{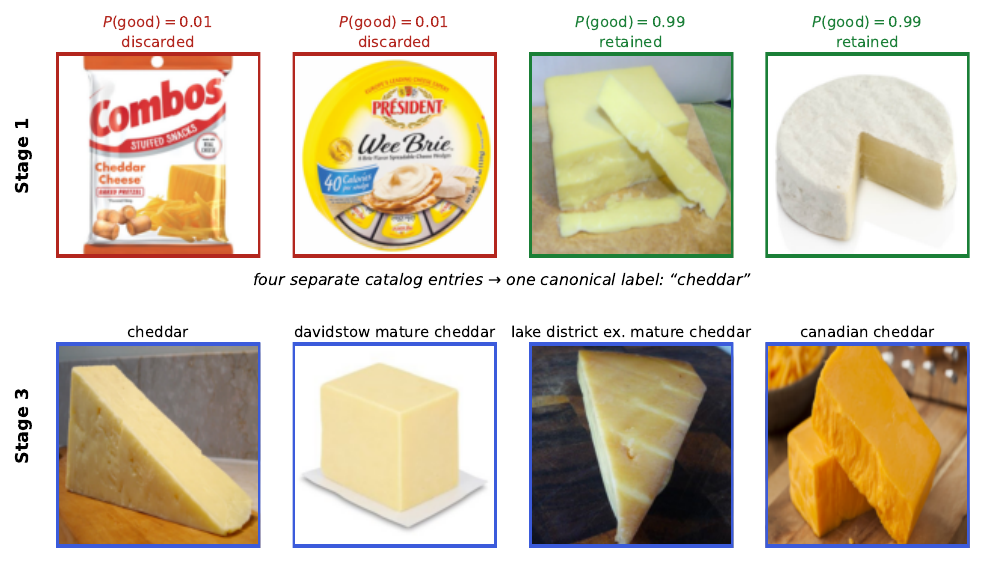}
\caption{The two defects. \textbf{Top:} the Stage-1 probe scores packaging
shots near zero and product shots near one. The leftmost image shows a packaged snack rather than
unpackaged cheese, returned by \emph{``cheddar cheese''}. \textbf{Bottom:} four
distinct photographs from folders the scrape treats as four classes and Stage~3
maps to \emph{cheddar}. This is why no per-image relabeling fixes a
defect of this kind.}
\label{fig:qualitative}
\end{figure}

Both defects of Section~\ref{sec:intro} are prevalent (Figure~\ref{fig:qualitative}), and folder sizes are highly uneven. The categories retaining at least twenty distinct clean images after Stage~1 filtering and deduplication form the recognition set, which remains both long-tailed and redundant (Section~\ref{sec:stage3}).

Canonicalization aligns cheese names to a tabular catalog scraped from \emph{cheese.com} and redistributed by the TidyTuesday project~\cite{tidytuesday2024cheese}, used as a fixed external reference. We redistribute no catalog text.

\begin{table}[!tb]
\caption{Archive and label-set statistics. The recognition subset exceeds the
distinct-photograph count because a duplicate group spanning several folders
contributes one image to each (Section~\ref{sec:dedup}).}
\label{tab:dataset}
\centering
\small
\setlength{\tabcolsep}{2pt}
\begin{tabular}{lr}
\toprule
Quantity & Value \\
\midrule
Folders (cheese names) & 1{,}824 \\
Total readable images & 165{,}769 \\
Images kept after packaging filter & 62{,}174 (37.5\%) \\
Distinct photographs after dedup & 33{,}583 (54.0\% of kept) \\
\midrule
Cheeses with $\geq 20$ distinct images & 998 \\
Distinct images/class, median / mean & 35 / 36.1 \\
\quad Recognition subset, images (cap 80) & 35{,}986 \\
\qquad train / val / test & 25{,}325 / 5{,}344 / 5{,}317 \\
\qquad distinct photograph groups, test & 5{,}086 \\
\midrule
Canonical classes after merge (Stage~3) & 844 \\
\bottomrule
\end{tabular}
\end{table}

\section{Method}
\label{sec:method}

\begin{figure*}[t]
\centering
\includegraphics[width=\textwidth]{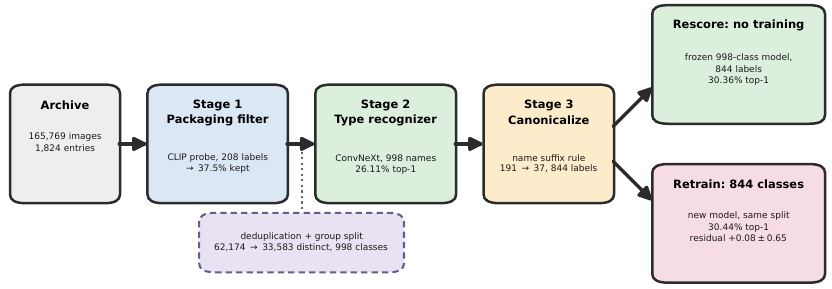}
\caption{The pipeline. Near-duplicate grouping reduces the filtered images to
$33{,}583$ distinct photographs before any split is drawn. Stage~3 changes the
label space, not the weights, so Stage~2's frozen model can be read out over it
without retraining. The two branches illustrate this comparison, decomposed in
Figure~\ref{fig:decomp}. Six-seed ConvNeXt-Small means (Table~\ref{tab:stage3}).}
\label{fig:pipeline}
\end{figure*}

Figure~\ref{fig:pipeline} summarizes the pipeline: one data-filtering stage and two treatments of the recognition labels. Stage~1 and deduplication determine which images enter the recognition set. Stage~2 recognizes the raw folder name. Stage~3 canonicalizes it and is evaluated both by rescoring Stage~2's frozen model and by retraining.

All CLIP features are $768$-dimensional $L_2$-normalized embeddings from ViT-L/14.

\begin{figure}[t]
\centering
\includegraphics[width=\columnwidth]{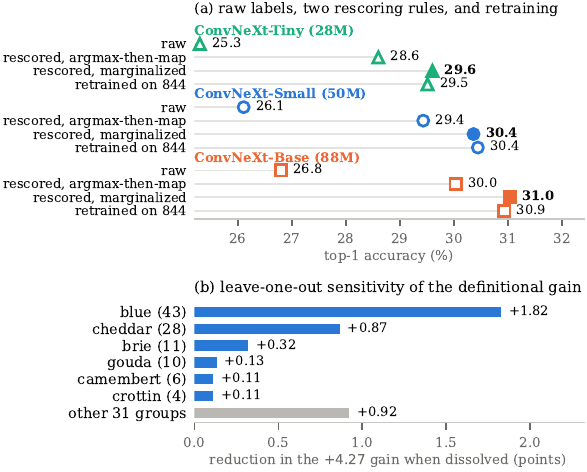}
\caption{\textbf{(a)} Raw-label evaluation, two canonical rescoring rules, and
canonical-label retraining, across three backbone sizes (six-seed means). The
filled marker shows the accuracy obtained from the frozen model's posterior over
canonical labels, without retraining. It matches the retrained model at
every size. \textbf{(b)} Leave-one-out
sensitivity: the gain is concentrated in \emph{blue}, which coarsens $43$ entries
into a broad type rather than collapsing aliases.}
\label{fig:decomp}
\end{figure}

\subsection{Stage 1: Packaging Filter}
\label{sec:stage1}

The filter is a binary classifier: does an image depict unpackaged cheese (``good'') or commercial packaging (``not\_good'')? We hand-label $208$ images ($105$ good, $103$ packaging) and fit a logistic regression on their CLIP features, with standardized features and balanced class weights. $5$-fold stratified cross-validation selects $C=0.01$. We then run the probe over all $165{,}769$ images and keep those scoring $P(\text{good})\!\geq\!0.5$. Discarding packaging fixes the task as recognition of the unpackaged product, not packaged-SKU recognition, where the label is on the wrapper and packaging is the signal.

\subsection{Deduplication and Leakage Control}
\label{sec:dedup}

Query-based scraping returns the same photograph more than once: the same JPEG under two filenames in one folder, and the same image at two resolutions under two cheese names. Splitting on filenames therefore leaks. Shuffling each folder's filenames into a $70/15/15$ train/validation/test split leaves $60.5\%$ of test images with a pixel-identical twin in training. The fine-tuned ConvNeXt-Small model of Section~\ref{sec:stage2} scores $43.8\%$ on these test images, compared with $28.7\%$ on the remainder: such a split measures memorization as much as recognition.

We therefore group the pool by image identity before splitting. Each image carries an MD5 hash and two $64$-bit perceptual hashes, a dHash over a $9{\times}8$ luminance gradient and an aHash against the image mean. Two images join a group when their MD5s match, or when \emph{both} hashes agree to within two bits. Candidate pairs come from banded lookup, so the $1.9$\,B pair space is never enumerated. The threshold is not tuned for effect: at two bits the largest group stays at its exact-duplicate value of $131$, whereas transitive chaining inflates it to $617$ at six bits. Across $400$ sampled merges missed by MD5 matching, the median pixel correlation is $0.9996$, consistent with re-encodings of the same photograph. The pairs a four-bit threshold would add have a median correlation of $0.63$. The sampled pairs give an estimated false-merge rate of $0.8\%$, and a false merge discards a photograph rather than creating leakage.

Two rules then build the split: each group contributes at most one image per folder, and every surviving member lands in the same split. No identified duplicate group therefore straddles train and test, even under two cheese names. Auditing candidate pairs added at four bits finds $49$ near-identical train--test pairs, so at most $49$ test images ($0.9\%$) carry an audited near-duplicate in training, against the $60.5\%$ a filename split leaves. The audit does not cover all possible pairs. The first rule deduplicates within folders but not across them, so one photograph can still appear in the test split under several names. The $5{,}317$ test images occupy $5{,}086$ groups, of which $103$ carry more than one canonical label ($304$ images, $5.7\%$). Conflicting labels on identical inputs impose unavoidable errors, though not one per image, since near-duplicate members may still receive different predictions. These repeated photographs also make image-level outcomes dependent (Section~\ref{sec:stats}).

The redundancy is severe (Table~\ref{tab:dataset}): $6{,}164$ duplicate groups span more than one catalog entry, and $5{,}566$ still span distinct labels under the recognition merge map, with unmapped entries kept separate. Holding the protocol constants fixed ($\geq\!20$ unique images per class, cap $80$, $70/15/15$), only $998$ of the $1{,}238$ entries that qualify based on filtered file counts still qualify based on distinct photographs.

\subsection{Stage 2: Fine-Grained Type Recognition}
\label{sec:stage2}

On the $998$-class deduplicated clean set, we compare three approaches in increasing order of cost.

\noindent\textbf{Zero-shot CLIP.} Each image is matched to the textual prompt \emph{``a photo of \{name\}''} for every cheese, and the argmax is the prediction. A single hand-written prompt confounds text grounding with prompt design. We therefore also report an eight-template ensemble, a variant using names stripped of the trailing \emph{``cheese''}, and a variant mapped to the canonical label space of Section~\ref{sec:stage3}.

\noindent\textbf{Linear probe.} We fit logistic regression ($C\!=\!1$) on standardized frozen CLIP features from the training split and evaluate it on the test split.

\noindent\textbf{Fine-tuned backbone.} We fine-tune ConvNeXt~\cite{liu2022convnext} from torchvision's \texttt{IMAGENET1K\_V1} weights end-to-end with AdamW~\cite{loshchilov2019adamw} ($\text{lr}=10^{-4}$, weight decay $0.05$), a cosine schedule, label smoothing $0.1$, mixed precision, and RandAugment~\cite{cubuk2020randaugment} over $224{\times}224$ random-resized crops. We report ConvNeXt-Tiny, -Small and -Base ($28$, $50$ and $88$\,M parameters), all trained for the full $25$ epochs on the identical deduplicated split (Section~\ref{sec:dedup}), a $3.1\times$ capacity range at a fixed epoch budget.

\subsection{Stage 3: Base-Type Canonicalization}
\label{sec:stage3}

This stage canonicalizes each entry to its base type by name alone. Whether a given group collects aliases of one product or distinct products of one type is not determined by the procedure, and Section~\ref{sec:stage3res} shows the distinction matters. Using \emph{cheese.com} names as anchors, the shortest catalog name that is a suffix of the folder name becomes the canonical label, so \emph{``orkney extra mature cheddar''} becomes \emph{cheddar}. A stop-list prevents marketing descriptors such as \emph{gold} or \emph{reserve} from becoming a base. For a folder whose nonempty, normalized token sequence, after removing trailing \emph{cheese} tokens, is $t=(t_1,\dots,t_n)$, with anchor set $\mathcal{A}$ and trailing-token frequencies $f(\cdot)$,
\begin{equation}
c(t)=
\begin{cases}
a(s^\star) & \text{if }\mathcal{S}(t)\neq\emptyset,\\
a(t_n) & \text{else if } t_n\notin\mathcal{M},\ f(t_n)\!\geq\!\tau,\\
t & \text{otherwise,}
\end{cases}
\label{eq:canon}
\end{equation}
where $\mathcal{S}(t)$ collects the suffixes of $t$ of at most three tokens that lie in $\mathcal{A}$, excluding single-token suffixes in the modifier stop-list $\mathcal{M}$, and $s^\star$ is its shortest element. The anchors are normalized catalog names, synonyms and alternative spellings of at least three characters from records with recognized type/rind metadata. Here $f(\cdot)$ counts trailing tokens over the $1{,}238$ eligible folders, $\tau\!=\!3$, and $a$ maps \emph{goats} to \emph{goat} and leaves other names unchanged. Restricted to the $998$-entry recognition set, Eq.~\ref{eq:canon} collapses $191$ entries into $37$ groups, reducing the label count to $844$. The largest groups are \emph{blue} ($43$ folders), \emph{cheddar} ($28$), \emph{brie} ($11$), \emph{gouda} ($10$) and \emph{goat} ($9$). The stop-list must also block trailing tokens that are not cheese types at all (designations of origin, milk animals, nationalities), which otherwise clear $\tau$ and pull unrelated products together, putting \emph{murazzano dop} and \emph{pecorino toscano dop} onto \emph{dop}. We then both (a) rescore the existing $998$-class model under the merged labels and (b) retrain on the $844$ classes, keeping the split identical.

What the unchanged model predicts in the merged space is a modeling choice, not a formality. Let $g(k)$ map raw class $k$ to its canonical label, and let $p_k(x)$ be the frozen model's softmax probability. We \emph{marginalize}:
\begin{equation}
q_j(x)=\sum_{k:g(k)=j}p_k(x),\quad
\hat y(x)=\arg\max_j q_j(x).
\label{eq:rescore}
\end{equation}
The five highest-scoring groups give top-5. Thus $q$ is the model's posterior over canonical labels, without retraining. A weaker alternative, \emph{argmax-then-map}, takes the $998$-way argmax and translates that one label, so a group whose mass is split across several raw entries can lose to a single entry outside the group. We report both, because argmax-then-map is what a curation report implicitly computes when it relabels predictions. Marginalizing is not merely the more generous rule: it is not monotone, losing $32$ test images the raw argmax got right while gaining $259$. The merge uses only string normalization and external catalog matching, with no visual features, model predictions or test-set information.

\section{Experimental Setup}
\label{sec:setup}

ConvNeXt test accuracies and paired comparisons use saved per-image predictions, not separate training-log evaluations. Differences are computed before rounding; accuracy differences are in percentage points. Unless otherwise stated, recognition experiments use the fixed deduplicated split of Section~\ref{sec:dedup} ($25{,}325$ / $5{,}344$ / $5{,}317$ images over $998$ categories, capped at $80$ per type, a cap no deduplicated class reaches) on one NVIDIA RTX~4060 (8\,GB). Stage-1 numbers are over the full archive. Frozen-feature baselines use the same splits as the fine-tuned models, so Table~\ref{tab:stage2} carries no protocol difference between rows.

The backbone, merged and class-balanced runs share the fine-tuning configuration of Section~\ref{sec:stage2}, so only the studied variable changes. Batch size is the one exception ($64$ for Tiny, $48$ for Small, $32$ for Base), set by the $8$\,GB memory ceiling rather than tuned. It is therefore not fixed along the capacity axis, and confounds every capacity step.

\section{Results}
\label{sec:results}

\subsection{Stage 1: Packaging Filter}
\label{sec:stage1res}

Over the full archive, the probe labels $62{,}174$ images ($37.5\%$) as real cheese and $103{,}595$ as packaging or otherwise unusable. The score distribution is bimodal but not cleanly separated, with $9.4\%$ of images in the ambiguous band between $0.4$ and $0.6$.

The filter gates every downstream result, so we evaluate it out-of-fold. On the $208$ hand-labeled images, $5$-fold cross-validation gives a precision of $90.4\%$ ($95\%$ Wilson interval $[83.2,94.7]\%$) and a recall of $89.5\%$ ($[82.2,94.0]\%$) for the ``good'' class. Two consequences matter downstream. The filter discards about one genuine image in ten, and precision on a near-balanced label set overstates deployment precision: assuming the cross-validation error rates transfer to the full archive, the observed $37.5\%$ retention implies a true prevalence near $34.8\%$ and precision near $83\%$ among retained images. Contamination after recognition-set selection is unmeasured.

\subsection{Stage 2: Recognition}
\label{sec:stage2res}

Table~\ref{tab:stage2} compares methods on the same held-out test split. Zero-shot CLIP reaches $5.4\%$, some $50\times$ the uniform random baseline ($998$-way, $0.10\%$) but far from usable, consistent with, but not establishing, limited text grounding of rare category names. A linear probe on the \emph{same} frozen features reaches $20.4\%$: discriminative structure is present in the frozen image representation even where the text encoder cannot address it. End-to-end fine-tuning adds a further $6$ points.

\noindent\textbf{Tested prompt variants do not close the gap.} An eight-template ensemble reaches $4.8\%$, slightly \emph{below} the single prompt. A cleaned-name prompt (\emph{``a photo of \{name\}, a type of cheese''}) gives $5.5\%$. Mapping ensemble predictions to canonical labels reaches $7.5\%$, using the maximum similarity within each group rather than the sum of probabilities. The spread across four protocols is under three points, so prompt design and label redundancy do not close the gap to the probe.

\noindent\textbf{Capacity buys little at this budget.} Top-1 moves $25.6 \rightarrow 26.3 \rightarrow 26.6\%$ on this split, and neither adjacent step separates over six seeds (Table~\ref{tab:stats}).

\begin{table}[!tb]
\caption{Stage 2: cheese-type recognition ($998$ classes). Every row is scored
on the same held-out test split ($N=5{,}317$) of the deduplicated $80$-capped
subset. The probe is fit on that split's $25{,}325$ training images. ConvNeXt-Small
remains the reference model for Sections~\ref{sec:stage3res} to~\ref{sec:stats},
with six-seed results in Table~\ref{tab:stage3}. ConvNeXt rows here use seed $42$. No row is bolded: no single capacity step
separates from its neighbor (Table~\ref{tab:stats}).}
\label{tab:stage2}
\centering
\small
\begin{tabular}{lcc}
\toprule
Method & top-1 & top-5 \\
\midrule
Uniform random baseline & 0.10\% & 0.50\% \\
Zero-shot CLIP (text-name match) & 5.4\% & 11.1\% \\
CLIP linear probe (frozen) & 20.4\% & 37.7\% \\
ConvNeXt-Tiny fine-tuned, 25 ep & 25.6\% & 44.1\% \\
ConvNeXt-Small fine-tuned, 25 ep & 26.3\% & 44.7\% \\
ConvNeXt-Base fine-tuned, 25 ep & 26.6\% & 45.4\% \\
\bottomrule
\end{tabular}
\end{table}

\begin{figure}[t]
\centering
\includegraphics[width=\columnwidth]{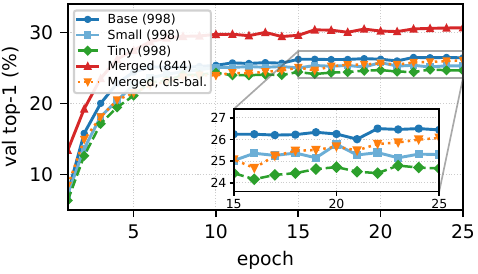}
\caption{Validation top-1 per epoch, seed $42$. The three raw-label backbones each
peak before epoch $25$ and their final accuracies differ by under two points, so
Table~\ref{tab:stage2} compares them at a fixed epoch budget. The two
$844$-class runs are plotted for reference. Inset: epochs $15$ to $25$.}
\label{fig:curves}
\end{figure}

Validation top-1 peaks at $24.8\%$ for Tiny, $25.8\%$ for Small and $26.5\%$ for Base, in each case before epoch $25$, and none has collapsed (Figure~\ref{fig:curves}). We therefore compare at a fixed $25$-epoch budget, which equalizes epochs but not compute or optimizer steps.

\subsection{Stage 3: Decomposing the Merge Gain}
\label{sec:stage3res}

Canonicalizing categories to their base type raises ConvNeXt-Small top-1 by $4.3$ points over six seeds, from $26.1\%$ to $30.4\%$ (Table~\ref{tab:stage3}). Writing $A_{\mathrm{raw}}$, $A_{\mathrm{marg}}$ and $A_{\mathrm{ret}}$ for raw, marginally rescored (Eq.~\ref{eq:rescore}) and retrained accuracies, we define $D=A_{\mathrm{marg}}-A_{\mathrm{raw}}$ and $R=A_{\mathrm{ret}}-A_{\mathrm{marg}}$, so $A_{\mathrm{ret}}-A_{\mathrm{raw}}=D+R$. We report six-seed means throughout and condition on the reference seed-$42$ run only for the paired per-image tests.

\noindent\textbf{Definitional ($+4.3$ points).} Scoring the \emph{unchanged} $998$-class model against the merged label space gives $30.36\%$ top-1 over six seeds, against $26.11\%$ raw, a $+4.25$-point gain (Table~\ref{tab:stage3}). Only the label space used to interpret the posterior changes, so predictions previously marked wrong for naming \emph{``mature cheddar''} when the label said \emph{``cheddar''} now count. This already matches the $30.44\%$ reached by retraining.

\noindent\textbf{Retraining residual ($+0.1$ points).} Over six seeds, retraining on the $844$ merged classes reaches $30.44\%$, indistinguishable from the rescored $30.36\%$ ($+0.08\pm0.65$; Table~\ref{tab:stage3}). On the reference seed-$42$ run the retrained model reaches $30.4\%$, slightly \emph{below} its rescored figure, and a paired McNemar test does not separate them ($b/c = 319/307$, $p=0.66$). The residual is indistinguishable from zero at all three capacities (Section~\ref{sec:stats}), so retraining on the consolidated catalog buys nothing we can resolve.

The rescoring rule carries this conclusion. Under argmax-then-map the same model reaches only $29.4\%$, leaving a $+1.0$-point residual, positive in all six seeds ($+1.01\pm0.58$, $p=0.008$), that reads as a genuine gain from retraining. A curation report that relabels predictions rather than re-reading the posterior will therefore credit about a point to retraining that the frozen model's own posterior already delivers.

\noindent\textbf{Is the definitional gain merely a smaller label space?}
Collapsing $998$ classes to $844$ makes the task easier by construction, so we compare it against nulls drawing $10{,}000$ random merges with the \emph{identical} $191\rightarrow37$ size profile, holding the model, predictions and split fixed: this class-set-matched null merges exactly the $191$ entries Stage~3 merges and shuffles only their grouping. Under marginal rescoring it gains $+0.68$ against the observed $+4.27$, which exceeds all $10{,}000$ draws (permutation $p<10^{-4}$).

The remaining comparisons are computed under argmax-then-map, where the corresponding observed gain is $+3.03$, ensuring a like-for-like comparison. A permute-all null randomizes which entries merge as well as how they group, and gains $+0.22$ against the class-set-matched $+0.42$. That splits the component three ways: class-count reduction $+0.22$, high-frequency entries $+0.19$, and grouping the \emph{specifically confusable} ones $+2.61$. A comparator merge built greedily from the model's own test confusion reaches $+3.71$, so exploiting the model's confusions does not do dramatically better. Being test-informed and greedy, it bounds nothing.

\noindent\textbf{What canonicalization merges, and why it matters.} The merge does two different things: it collapses several names for one product (an \emph{alias collapse}) and it coarsens distinct products into a broader type (a \emph{type coarsening}). We hand-label each of the $37$ merged groups as one or the other and split the gain accordingly. It is overwhelmingly the second kind: the $29$ type-coarsening groups supply $+4.01$ points ($94\%$ of the definitional gain), while the $8$ alias collapses supply only $+0.28$ ($7\%$). Leave-one-out sensitivities localize it further: \emph{blue}, which gathers $43$ distinct blue cheeses into one type, accounts for $1.82$ points ($43\%$) on its own, \emph{cheddar} ($28$ entries) for $0.87$, and the five largest groups for $74\%$ (Figure~\ref{fig:decomp}b). These compete under marginalization rather than partition the gain, so they need not sum to the total ($4.29$ against $4.27$). This reading is also the paper's practical diagnostic. The alias-versus-coarsening split separates the part of a curation gain that is duplicate-label correction from the part that is merely an easier task, and on this archive the gain is almost all the latter: a curation step that merged genuine aliases would show the opposite profile.

\noindent\textbf{The definitional component is stable across recognizers.}
The three backbones agree on this component to within $0.1$ points across a $3.1\times$ parameter range, and a frozen CLIP probe gives $+3.76$ (Table~\ref{tab:stage3}); a stricter $0.7$ filter raises it to $+5.09$ on a non-comparable split. Two cautions bound this. The probe's lower value shows the component still depends on feature quality when features are weak enough. And it is a \emph{net} change under a new decision rule, not a count of recovered errors: marginalizing can also lose images the raw argmax got right, when a merged group's pooled mass overtakes the correct singleton. The gains under both rules are bounded above by the raw error rate, so we cannot extrapolate beyond the $20$ to $27\%$ band measured here.

\noindent\textbf{No detectable interaction with capacity.} If a redundant label space suppressed the return on capacity, the capacity step would be larger once the catalog is canonical. Over the full $3.1\times$ range it is not: paired within six seeds, Tiny-to-Base is worth $+1.50\pm0.57$ raw ($p=0.001$) and $+1.42\pm0.44$ canonical ($p=0.0005$). Comparing two such estimates is not itself a test. We therefore form the interaction within each seed, $I_s=(\text{Base}-\text{Tiny})_{\text{canonical}}-(\text{Base}-\text{Tiny})_{\text{raw}}$. It is $-0.08$, with a $95\%$ interval of $[-0.79,+0.63]$. That interval still admits effects large relative to the $1.50$-point capacity benefit itself. A single seed suggests otherwise. At seed $42$ the Small-to-Base step is $+0.21$ raw and $+1.02$ canonical, the pattern one would report from one run.

\noindent\textbf{Class-balanced loss.} Inverse-frequency weighted cross-entropy on the merged task (one seed) reduces micro top-1 from $30.4\%$ to $26.3\%$ (Table~\ref{tab:stats}) and raises macro-recall from $25.3\%$ to $25.8\%$. It buys a tail gain ($17.7\%\rightarrow20.4\%$ in the rarest tertile by training count) with a larger head loss ($37.4\%\rightarrow28.8\%$). Micro accuracy follows the head classes that dominate the test distribution.

\begin{table}[!tb]
\caption{Decomposing the merge gain: six-seed mean top-1 for the three
ConvNeXt backbones. Accuracies are percentages; components are percentage points with six-seed standard deviations.
Each row's own raw model is rescored under the same canonical map by the marginal
rule. The last two rows are single runs, and the $0.7$-filter row rebuilds the
split, so only its decomposition is comparable.}
\label{tab:stage3}
\centering
\small
\setlength{\tabcolsep}{2.2pt}
\begin{tabular}{lccccc}
\toprule
Model & raw & marg. & retr. & defin. & resid. \\
\midrule
Tiny \ ($28$\,M)           & 25.30 & 29.60 & 29.51 & $+4.30$\tiny{$\pm.24$} & $-0.10$\tiny{$\pm.53$} \\
Small ($50$\,M)            & 26.11 & 30.36 & 30.44 & $+4.25$\tiny{$\pm.09$} & $+0.08$\tiny{$\pm.65$} \\
Base \ ($88$\,M)           & 26.80 & 31.03 & 30.93 & $+4.23$\tiny{$\pm.20$} & $-0.10$\tiny{$\pm.65$} \\
\midrule
CLIP probe, frozen         & 20.44 & 24.21 & 23.34 & $+3.76$ & $-0.87$ \\
Small, filter $0.7$        & 25.66 & 30.75 & 30.70 & $+5.09$ & $-0.05$ \\
\bottomrule
\end{tabular}

\end{table}

\subsection{Statistical Reliability}
\label{sec:stats}

We assess uncertainty in two ways: paired comparisons on the reference run, and repeated training with the split held fixed.

A top-1 accuracy near $26\%$ on this test split has an image-level $95\%$ Wilson half-width of about $1.2$ points. Individual intervals do not give the interval of a paired difference. Because every model is scored on the same images, we compare their paired outcomes with exact McNemar tests~\cite{dietterich1998approximate} (Table~\ref{tab:stats}). Both the Wilson intervals and McNemar tests treat test images as independent, which the residual cross-folder duplication of Section~\ref{sec:dedup} violates: pairing is appropriate, since all runs are scored on the identical $5{,}317$ images, but those images are not independent draws. We therefore rest substantive conclusions on group-aware intervals rather than on these $p$-values. Percentile intervals from $5{,}000$ bootstrap resamples of the $5{,}086$ photograph groups give $+4.27$ $[+3.61,+4.91]$ definitional and $-0.23$ $[-1.09,+0.68]$ residual, so the dependence widens intervals without changing either conclusion. Restricting to the $5{,}013$ images whose group carries one canonical label gives $+4.11$ and $-0.24$.

\begin{table}[!tb]
\caption{Exact McNemar tests on paired per-image outcomes, shared test split
($N=5{,}317$), with the discordant counts used in each test: $b$ correct
before and wrong after, $c$ wrong before and correct after.}
\label{tab:stats}
\centering
\small
\setlength{\tabcolsep}{2.6pt}
\begin{tabular}{lccc}
\toprule
Comparison & $\Delta$ & $b$\,/\,$c$ & McNemar $p$ \\
\midrule
Tiny $\rightarrow$ Small (capacity)       & $+0.8$ & 322\,/\,363 & $0.13$ \\
Small $\rightarrow$ Base (capacity)        & $+0.2$ & 317\,/\,328 & $0.69$ \\
Tiny $\rightarrow$ Base (capacity)         & $+1.0$ & 302\,/\,354 & $0.046$ \\
Small $\rightarrow$ Base, merged labels    & $+1.0$ & 279\,/\,333 & $0.03$ \\
Merged $\rightarrow$ merged + class-bal.  & $-4.1$ & 429\,/\,213 & $1\!\times\!10^{-17}$ \\
\midrule
Raw $\rightarrow$ rescored, marginal      & $+4.3$ & \ \ 32\,/\,259 & $3\!\times\!10^{-45}$ \\
\quad argmax-then-map $\rightarrow$ marginal & $+1.2$ & \ \ 32\,/\,\ 98 & $6\!\times\!10^{-9}$ \\
Rescored $\rightarrow$ retrained (resid.) & $-0.2$ & 319\,/\,307 & $0.66$ \\
\bottomrule
\end{tabular}
\end{table}

\noindent\textbf{Seed replication.} The paired tests condition on one run, so we retrain every backbone in both label spaces under six seeds ($42$ and $1$ to $5$) with the split held fixed. We test within-seed differences with two-sided one-sample $t$ tests with $5$ degrees of freedom. Table~\ref{tab:stage3} gives the resulting means and the decomposition repeated within each seed. The $0.7$-threshold and class-balanced runs remain single-seed. The residual is indistinguishable from zero at all three capacities ($p=0.67$, $0.78$ and $0.71$, positive in three of six seeds in each case). Paired by seed, Small exceeds Tiny by only $+0.81$ points, less than the $0.88$ spread of the per-seed differences themselves ($t=2.27$, $p=0.073$). Neither the split nor the seeds resolve that step, while Tiny-to-Base does. Under argmax-then-map the same runs give residuals of $+0.91\pm0.62$, $+1.01\pm0.58$ and $+0.90\pm0.65$. That component exists only relative to the weaker scoring rule, and it does not grow with capacity either.

\section{Discussion}
\label{sec:discussion}

\noindent\textbf{The gain is real, and it is bookkeeping.} Sometimes the base type is the required output, as in catalog normalization or product-type retrieval over unpackaged goods. There the $+4.3$ definitional points are delivered accuracy. Where the entries are genuinely different purchasable products, they are not, since predicting \emph{cheddar} does not identify \emph{mature cheddar}. Which applies is a per-group question, and the gain is concentrated in exactly the groups where it is least clear. But the gain is available as soon as the merge map is defined, from a model already trained. The consolidation changed what the system is asked to say, and parity in accuracy is not evidence that the learned features are unchanged. Absolute top-1 also stays near $30\%$ even after consolidation, an $844$-way fine-grained problem over long-tailed web photographs; this rules out the checkout and shelf-audit uses that nominally motivate retail recognition, which must name a packaged SKU and whose signal our packaging filter (Section~\ref{sec:stage1}) deliberately removes. What the pipeline supports is the catalog-side work above, where the decomposition, not the absolute number, is the contribution.

\noindent\textbf{Grounding, hierarchy and deployment.} Neither prompt design nor label redundancy closes the zero-shot gap (Section~\ref{sec:stage2res}), and we did not vary the text tower, so its ability to represent rare proper nouns remains untested; that is a reason not to assume open-vocabulary recognition will absorb catalog maintenance. A hierarchy adopted as evaluation ground truth deserves the same near-duplicate audit and rescoring control as this flat label space. Top-5 ($50.1\%$) favors a ranked shortlist over one guess, but the correct label is missing from about half the shortlists, motivating an assisted-identification study rather than a checkout deployment.

\section{Limitations}
\label{sec:limitations}

\noindent\textbf{One split, one budget.} Six seeds characterize training stochasticity, but all runs share one partition, so split-to-split variation is unmeasured. Capacity is explored only at $224{\times}224$ for $25$ epochs, so every capacity statement is \emph{at this budget}.

\noindent\textbf{The Stage-1 filter is thinly supervised, and it selects.} Out-of-fold precision and recall are near $90\%$. But they rest on $208$ labels cross-validated over images rather than folders, and the regularization strength was chosen on those same folds rather than in a nested loop. Both effects make them optimistic. The archive-level estimate of Section~\ref{sec:stage1res} implies substantial residual packaging. It also thins the catalog unevenly (mean per-folder retention $37.3\%$, standard deviation $24.1$ percentage points), only weakly associated with folder size ($\rho=0.057$, $p=0.014$), though the $191$ merged entries retain more than the $807$ untouched ones ($56.8\%$ vs $53.2\%$, $p=0.008$).

\noindent\textbf{A null, not a proof of absence.} For ConvNeXt-Small, the retraining residual is $+0.08\pm0.65$ over six seeds, with a $95\%$ confidence interval of $[-0.60,+0.76]$ points, so an effect of a few tenths would not be detected. We claim only that retraining buys nothing at the scale of the reported curation gain, and we set no equivalence margin. The quantity also mixes representation with head fitting, calibration and optimization, hence \emph{retraining residual} rather than a claim about features. Separately, class sizes are limited per raw entry, before the merge, so canonical \emph{blue} holds $1{,}140$ training images against a per-class maximum of $50$ in the raw model. Capping the training split at $50$ per canonical class costs $4.4$ points (retrained canonical mean $30.6\rightarrow26.2$ over the same three seeds). But that control alters class composition as well as training volume, since the $31$ classes it touches carry $19\%$ of the test set. It therefore neither isolates the volume advantage nor bounds its effect on the residual.

\noindent\textbf{The merge map is name-based, not image-verified.} We hand-classify the $37$ merged groups as alias collapses or type coarsenings (Section~\ref{sec:stage3res}), but the merges themselves are formed by name rule and are not inspected image by image, so a group we label a coarsening could still contain near-identical products, or the reverse. The definitional component should therefore be read as the gain from scoring in this canonical space, with the alias-versus-coarsening split as its interpretation, not as an exact measure of catalog duplication. Coverage is the sharper limit. All $1{,}824$ folders clear $20$ raw images, but only $998$ do after filtering and deduplication. That leaves $45\%$ of the archive's named categories outside the label space, most lost to the filter. The tail measured here is not the tail a deployment would face.

\section{Conclusion}
\label{sec:conclusion}

We separate a catalog-curation gain into the part reflecting a changed task and the part that could reflect a better model. On this archive the second yields no statistically resolved improvement: scoring the unchanged model against the merged labels already recovers the full $+4.3$-point gain, and retraining adds nothing we can resolve over six seeds at three capacities spanning $3.1\times$ in parameters. That gain is also overwhelmingly a category coarsening ($94\%$) rather than alias correction, so the same decomposition both prices the retraining residual and tells a curator which kind of merge produced the gain. A curation gain that changes the label space should be reported relative to a frozen-model baseline rescored in that same space; this control costs one evaluation pass.

{ \small \bibliographystyle{ieeenat_fullname} \bibliography{main} }

\end{document}